\documentclass[conference]{IEEEtran}
\IEEEoverridecommandlockouts
\pdfoutput=1
\usepackage{cite}
\usepackage{amsmath,amssymb,amsfonts}
\usepackage{algorithmic}
\usepackage{graphicx}
\usepackage{textcomp}
\usepackage{xcolor}
\usepackage[noblocks]{authblk}
\usepackage{etoolbox}
\usepackage{multirow}
\usepackage{graphicx}

\usepackage{hyperref}
\def\BibTeX{{\rm B\kern-.05em{\sc i\kern-.025em b}\kern-.08em
    T\kern-.1667em\lower.7ex\hbox{E}\kern-.125emX}}
\begin{document}
\makeatletter
\newcommand{\printfnsymbol}[1]{%
  \textsuperscript{\@fnsymbol{#1}}%
}
\makeatother
\title{Deep Bi-Dense Networks for Image Super-Resolution}

\author[1]{Yucheng Wang\printfnsymbol{1}\thanks{\printfnsymbol{1} equal contribution}}
\author[2]{Jialiang Shen\printfnsymbol{1}}
\author[2]{Jian Zhang}
\affil[1]{Intelligent Driving Group, Baidu Inc}
\affil[2]{Multimedia and Data Analytics Lab, University of Technology Sydney\authorcr

        \texttt{
            \href{mailto:wangyucheng@baidu.com}{\nolinkurl{wangyucheng@baidu.com}}, \href{mailto:jialiang.shen@student.uts.edu.au}{\nolinkurl{jialiang.shen@student.uts.edu.au}}, \href{mailto:jian.zhang@uts.edu.au}{\nolinkurl{jian.zhang@uts.edu.au}}
        }

}

%\and
%\IEEEauthorblockN{4\textsuperscript{th} Given Name Surname}
%\IEEEauthorblockA{\textit{dept. name of organization (of Aff.)} \\
%\textit{name of organization (of Aff.)}\\
%City, Country \\
%email address}
%\and
%\IEEEauthorblockN{5\textsuperscript{th} Given Name Surname}
%\IEEEauthorblockA{\textit{dept. name of organization (of Aff.)} \\
%\textit{name of organization (of Aff.)}\\
%City, Country \\
%email address}
%\and
%\IEEEauthorblockN{6\textsuperscript{th} Given Name Surname}
%\IEEEauthorblockA{\textit{dept. name of organization (of Aff.)} \\
%\textit{name of organization (of Aff.)}\\
%City, Country \\
%email address}
%}

\maketitle

\begin{abstract}
This paper proposes Deep Bi-Dense Networks (DBDN) for single image super-resolution.
Our approach extends previous intra-block dense connection approaches by including novel inter-block dense connections. In this way, feature information propagates from a single dense block to all subsequent blocks, instead of to a single successor.

To build a DBDN, we firstly construct intra-dense blocks, which extract and compress abundant local features via densely connected convolutional layers and compression layers for further feature learning.
Then, we use an inter-block dense net to connect intra-dense blocks, which allow each intra-dense block propagates its own local features to all successors.
Additionally, our bi-dense construction connects each block to the output, alleviating the vanishing gradient problems in training.
The evaluation of our proposed method on five benchmark data sets shows that our DBDN outperforms the state of the art in SISR with a moderate number of network parameters.

\end{abstract}
\begin{IEEEkeywords}
Image super-resolution, CNN, Dense connection
\end{IEEEkeywords}

\section{Introduction}
The process of reconstructing high-resolution (HR) images from their low-resolution images (LR) is referred as super-resolution (SR).
SR has a wide range of applications, such as medical imaging \cite{b4}, satellite imaging \cite{b23}, and security surveillance \cite{b21}.
Image SR methods can be roughly divided into three categories:
(1) interpolation-based methods, such as nearest neighbouring interpolation, bilinear interpolation, and bicubic interpolation \cite{b3}, which assume that the intensity at a position on the HR image can be interpolated from its neighbouring pixels on the corresponding LR image.
(2) Optimisation-based methods \cite{b12} often model the problem by finding the optimum of an energy function, which consists of a data observation term from LR image and a regularisation term from various assumptions and hypotheses.
(3) Learning-based methods, directly learn the mapping function from the LR image to the HR image \cite{b24}.

Driven by the emergence of large-scale data sets and fast development of computation power, learning-based methods especially using deep neural networks \cite{b1}\cite{b5}\cite{b6}\cite{b7} have proven effective for image SR.
However, as CNNs become increasingly deep, a new problem emerges: as the gradient passes through many layers, it can vanish by the time it reaches the beginning of the network.
Many recent publications, such as VDSR and DRCN \cite{b6}\cite{b7} address this or related problems.
These approaches vary in network topology and training procedure, but they share a common characteristic:
\begin{figure}[t]
\includegraphics[width=0.45\textwidth,height=0.20\textheight]{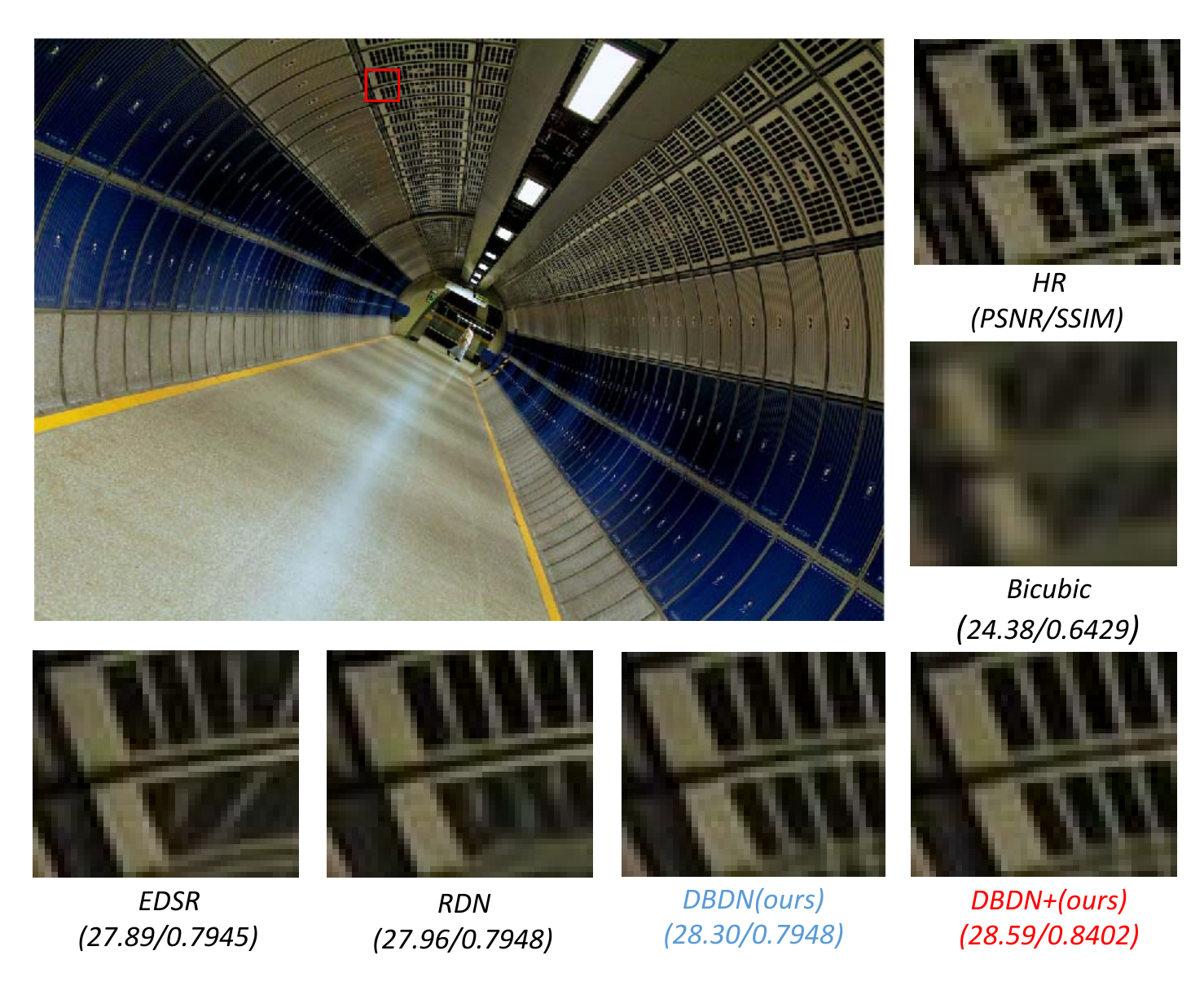}
\caption{Super-resolution result of our method on $4\times$ enlargement compared with existing algorithms}
\label{fig}
\end{figure}
they create short skip connections from early to later layers.
Among these methods, the most well-known one is SRResNet \cite{b8}.
To ease optimization, SRResNet adds skip connections that bypass the non-linear transformations with identity functions.
Lim et al. optimize the residual network in EDSR \cite{b14} for image SR by removing unnecessary modules and achieve excellent performance on the benchmark datasets.
Tai et al. propose multiple memory-efficient residual network structures by using recursive-supervision, such as DRRN \cite{b5} and MemNet \cite{b25}.
Although residual networks have proven to facilitate gradient flow in training, performance is limited by constraining feature reuse to one early layer, rather than all of the layers in a block.

In order to maximize the use of feature layers in the block,  Dense Convolutional Network (DenseNet) \cite{b26} was proposed to archive higher performance than ResNet \cite{b8}. In contrast to ResNets, DenseNet has two modifications.
First, instead of summation, DenseNet concatenates layers to preserve features for subsequent reuse.
Second, instead of connecting two layers in the residual block, DenseNet\cite{b26} connects all the layers in the block.
This modification helps DenseNet to achieve better performance with fewer parameters than ResNet.
This result indicates that feature learned at the early layers of the network matter to the task of SR, Therefore, by making more use of information in the feature layers, performance can be boosted.
Inspired by the success of DenseNet \cite{b26} in image classification, Tong et al. \cite{b15} propose SRDenseNet for image SR.
They remove the pooling layers of DenseNet\cite{b26} to make it more suitable for image SR.
Then they add skip connections between the blocks to mitigate the training of SRDenseNet\cite{b15}.
Zhang et al. \cite{b13} also introduce the dense blocks in RDN.
Compared with SRDenseNet \cite{b15}, RDN \cite{b13} uses larger growth rates and deep supervision to further improving the performance.
However, all of these methods only reuse the local feature layers in the dense block, and pass on the block information to one neighbour block for feature learning.
Haris et al. \cite{b32} propose DBPN which is constructed by iterative up- and down- sampling blocks. All the early upsampling blocks are concatenated as the input for the next downsampling block, or all the early downsampling blocks are concatenated as the input for the next upsampling block.
Therefore, each block information can't be reused by all the other blocks, which restricts the block information flow during propagation.

In order to make better use of block information, we propose a deep bi-dense network (DBDN) to enhance block information flow by introducing a novel inter-block dense connection to the network.
DBDN is built by intra-dense blocks and an inter-block dense net. Each layer connects to all the other layers in the intra-dense block and the output of the intra-dense block is the compressed concatenation of each layer features in the block.
Then, we use an inter-block dense net to connect these blocks. So each intra-dense block propagates its own local features to all successors.
We reconstruct the SR image through concatenating all the blocks' outputs, forming a direct supervision pattern. For this reason, each block has direct access to the gradient from the loss function and ground truth. Due to the bi-dense architecture in the network, our DBDN outperforms the state-of-art methods on the benchmark datasets.

In summary, our work provides the following contributions:
\subsubsection{Bi-dense architecture }
We propose a novel model called DBDN for image SR tasks. The model not only reuses local feature layers in the dense block, but also reuses the block information in the network to archive excellent performance with moderate parameter numbers.
\subsubsection{Intra-block dense connection}
We propose a compact intra-dense block where each layer is connected to every other layer to learn local features. Then, in order to preserve feature information and keep the model compact, we use the compressed concatenation of all layers' output in the block as the block output.
\subsubsection{Inter-block dense connection}
We introduce an inter-block dense net for high-level feature learning. Since the features learned in the early blocks of the network matter to the task of SR, we use inter-block dense connection to allow all of the early block information to be reused to learn the later block features. Furthermore, as all blocks have access to the output, the ground truth will directly supervise each block. The direct supervision alleviates the effect of vanishing/exploding gradients, and further improves the directness and transparency of the network.
\begin{figure*}[t]
\begin{center}
\includegraphics[width=0.85\textwidth,height=0.25\textheight]{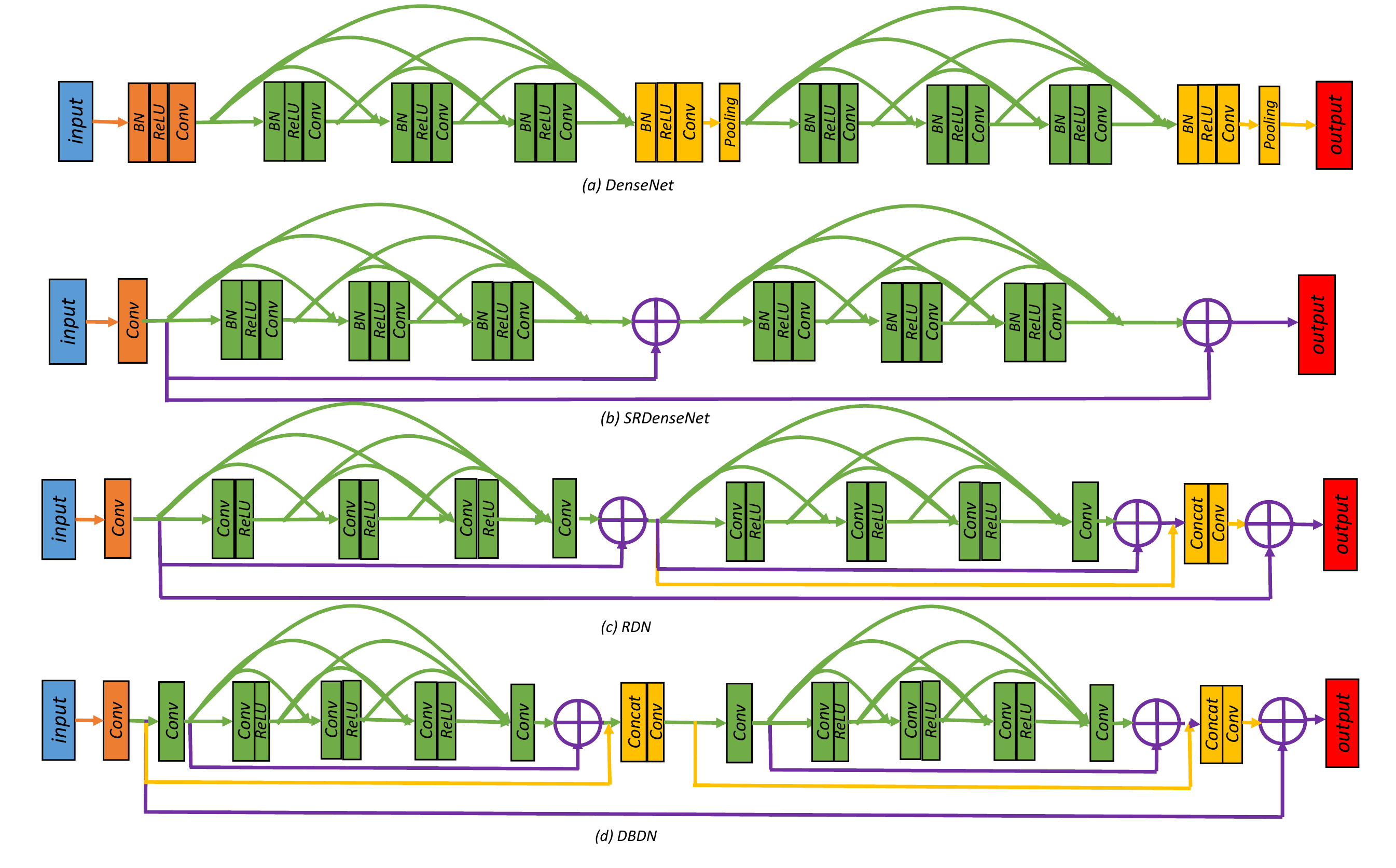}
\caption{Simplified structure of (a) DenseNet \cite{b26}. The green lines and layers denote the connections and layers in the dense block, and the yellow layers denote the transition and pooling layer. (b) SRDenseNet \cite{b15}. The green layers are the same dense block structures as those in DenseNet. The purple lines and element-wise $\bigoplus$ refer to the skip conection. (c) RDN \cite{b13}. The yellow lines denote concatenating the blocks output for reconstruction and the green layers are residual dense blocks. (d) DBDN. The yellow lines and layers denote the inter-block dense connection and the green layers are the intra dense blocks. The output goes into upsampling reconstruction layers. The orange layers after input are the feature extraction layers in all models. }
\end{center}
\label{compare}
\end{figure*}
\section{Related Works}
Since we overview learning-based methods in the introduction section, in this section we will focus on the three works that are most related to ours: DenseNet \cite{b26}, SRDenseNet \cite{b15}, and RDN \cite{b13}. To not lose generality, only two blocks are shown in Fig. 2. to describe these models.
\subsection{DenseNet}
The main idea of DenseNet \cite{b26} is to connect each layer to every other layer in a feed-forward network, in order to alleviate the vanishing gradient problems, strengthen feature propagation, and encourage feature reuse in the training of very deep networks. To implement the idea, the author proposes a dense connectivity mechanism as shown in Fig. 2(a), where each layer obtains additional inputs from all the preceding layers and passes on its own feature maps to all subsequent layers.
Denoting the input of each block as $x_{0}$, and the $\ell_{th}$ layer in the dense block as $H_{\ell}$, then the output of the $\ell_{th}$ layer $x_{\ell}$ in the block can be described as:
\begin{equation}
x_{\ell}=H_{\ell}([x_{0},x_{1},...,x_{\ell-1}])       \label{1}
\end{equation}
Here $[x_{0},x_{1},...,x_{\ell-1}]$ refers to the concatenation of layers produced before the $\ell_{th}$ layer.
$H_{\ell}(.)$ is a composite function of three consecutive operations: Batch Normalization (BN) \cite{b11}, followed by a rectified linear unit (ReLU) \cite{b12}, and a $3\times3$ Convolution (Conv).
Supposing the $b_{th}$ dense block structure has $\ell$ layers, then the output of the $b_{th}$ dense block is:
\begin{equation}
 B_{b}=[x_{0},x_{1},...,x_{\ell-1},x_{\ell}]        \label{2}
\end{equation}

Transition layers and pooling layers connect such dense blocks into line and change the block feature dimensions to construct the network. Although the dense connectivity in the dense blocks fulfills the idea of connecting each layer to all the other layers in the blocks. Stacking the dense blocks to construct DenseNet \cite{b26}, restricts the block information flow. Furthermore, the dense block in DenseNet \cite{b26} is designed for image recognition task, but it is not suitable for image SR and therefore needs to be adjusted for SR design.
\subsection{SRDenseNet}
Observing the efficiency of the dense block in DenseNet\cite{b26}, SRDenseNet \cite{b15} first introduces the dense connectivity into image SR. As shown in Fig. 2(b), the dense block in SRDenseNet \cite{b15} has the same structure as DenseNet \cite{b26}.
However, SRDenseNet \cite{b15} removes the transition layers and pooling layers. To keep the image detail information, removing the pooling layers is necessary in SR tasks. However, removing the transition layers will limit the setting of the feature layer channels. Then SRDenseNet adds skip connection between each block output and the network input, which helps to improve the network performance compared to only stacking the dense blocks.
Therefore the output of the $b_{th}$ dense block can be defined as
\begin{equation}
 B_{b}=[x_{0},x_{1},...,x_{l-1},x_{l}]+x      \label{3}
\end{equation}
Here x is the low level extracted features after convolutional layer.
Furthermore, SRDenseNet \cite{b15} has proven that adding a reasonable amount of skip connection between the blocks in the DenseNet \cite{b26} can improve the SR reconstruction performance. For this reason, ensuring more block information flow will potentially boost the image reconstruction accuracy.
\subsection{RDN}
Compared with SRDenseNet \cite{b15} and DenseNet \cite{b26}, RDN \cite{b13} is more suitable for SR tasks.
There are several notes for RDN:
(1) Unlike SRDenseNet \cite{b15} and DenseNet \cite{b26} that use a composite function of three consecutive operations $H_{\ell}(.)$ : BN, ReLU and Conv for each layer in dense block, inspired by EDSR \cite{b14}, RDN removes the BN module in the layer. The adjusted dense layers achieve better performance with fewer parameters compared with the original dense layers design in the SRDenseNet and DenseNet.
(2) RDN \cite{b13} uses a larger growth rate in covolutional layers, which allows it to learn more features compared with SRDenseNet \cite{b15}. In SRDenseNet, the growth rate is 16, but RDN has 64 growth rates. The wider network design benefits RDN and helps it to achieve higher performance.
(3) They concatenate all the block output for reconstruction, which alleviates the gradient vanishing problems in the deep network structure.
Therefore, the output of the $b_{th}$ residual dense block structure $B_{b}$ can be defined as
\begin{equation}
B_{b}=F_{b}([x_{0},x_{1},...,x_{l-1},x_{l}])+x_{0}   \label{4}
\end{equation}
, Here $F_{b}$ denotes the feature fusion function in the $b_{th}$ residual dense block. And The output for the reconstruction layer R can be described as
\begin{equation}
R=F([B_{0},B_{1},...,B_{b-1},B_{b}])+B_{0}      \label{5}
\end{equation}
, Here $F$ denotes the global feature fusion function in the network.
Although RDN concatenates all of the blocks' output for reconstruction at the end of the network, they don't reuse the block information by dense connection to learn features. Since the block only receives one early block information, and can't reuse all the early blocks information. It may lead to the block information and gradient loss during propagation. In this reason, we propose DBDN to reinforce the block information flow in the network.

\section{ Methods}
\begin{figure*}[t]
\begin{center}
\includegraphics[width=0.96\textwidth,height=0.20\textheight]{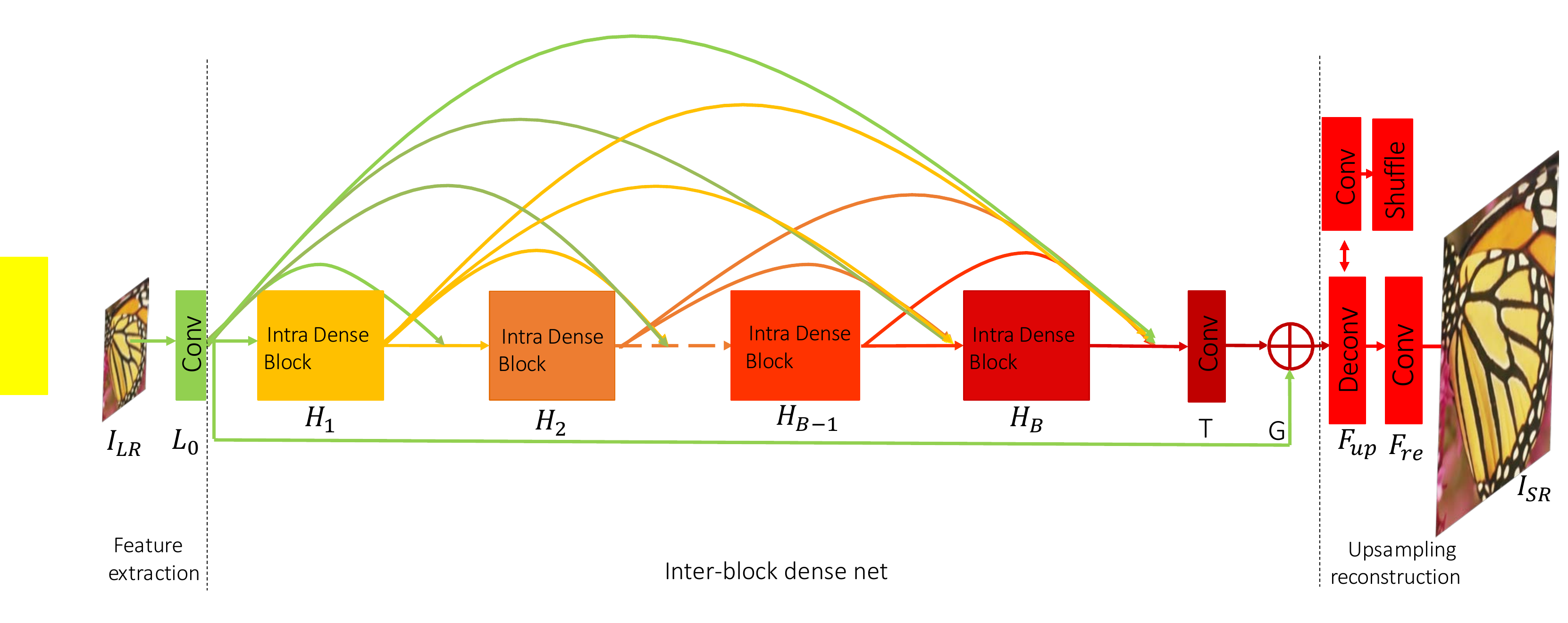}
\caption{Network structure}
\end{center}
\label{inter}
\end{figure*}

\subsection{Network structure}
The network is used to estimate a HR image $I_{SR}$ from a given LR image $I_{LR}$ , which is downscaled from the corresponding original HR image $I_{HR}$.
Our proposed network structure is outlined in Fig. 3.
The network can be divided into several sub-network structures: feature extraction layers for learning the low-level features, inter-block dense net for learning the high-level features, and upsampling reconstruction layers to learn upsampling features and to produce the target HR image. Here, $Conv(f,n)$ denotes one convolutional layer, where f is the filter size and n is the number of filters.

We use one convolutional layer to extract the low-level features $L^{0}$ from input images $I_{LR}$:
\begin{equation}
L^{0}=Conv(3,n_{r})(I_{LR})\label{6}
\end{equation}
Here the feature extraction layer produces $n_{r}$ channel features for further feature learning and global skip connection.

Following the initial feature extraction layer is the inter-block dense net. The Inter-block dense net is the main component to learn features for SR, where intra-dense blocks are densely connected to learn high-level features. Then we further use the compression layer to compress the output features and skip connection between the inter-block dense net input and output to mitigate the training of network. The output of the inter-block dense net G can be obtained by
\begin{equation}
G=H_{Inter}(L^{0})\label{7}
\end{equation}
Here $H_{Inter}$ denotes the function of the inter-block dense net. More details about the inter-block dense net will be given in the next subsection.

For upsampling layers, there are two different types of upsampling sub-networks, which are illustrated in Fig. 4.
One is called the deconvolution \cite{b20} layer and is an inverse operation of a convolution layer, which can learn diverse upsampling filters that work jointly for predicting the HR images. The other one is called the sub-pixel convolutional layer \cite{b19}, where the $I_{SR}$ image is achieved by the period shuffling features produced by the previous convolution layer. In order to have a fair comparison with EDSR \cite{b14} and RDN \cite{b13}, we propose the baseline models with deconvolution layer for upsampling as DBDN and the model with sub-pixel upsampling layer as DBDN plus (DBDN+).
Finally the target HR image is formulated as:
\begin{equation}
I_{SR}=Conv(3,3)(F_{up}(G)) \label{8}
\end{equation}
Here, $F_{up}$ denotes the upsampling layer and the reconstruction layer is a three-channel output convolutional layer.

%Given a training set consisting of HR image examples $I_{n}^{HR}, n= 1...N$, we generate the corresponding LR images $I_{n}^{LR}, n= 1...N $. Inspired by EDSR \cite{b14}, instead of calculating the pixel-wise mean squared error between the $I_{n}^{HR}$ and $I_{n}^{LR}$ as an objective function to train the network, we use L1 loss to train the network.
%So the loss function can be defined as:
%\begin{equation}
%l(\Theta)=\frac{1}{N}\sum_{n=1}^{N}\|F(I_{n}^{LR},\Theta)-I_{n}^{HR}\| \label{16}
%\end{equation}
%Here $\Theta$ denotes the network parameters and F means the function of the network. We minimize the loss function to optimize the network.
\subsection{Inter-block dense net}
Now we present more details about our proposed inter-block dense net. As shown in Fig. 3, Our inter-block dense net consists of inter-block dense connections, a global compression layer and a global skip connection .
\subsubsection{Inter-block dense connection}
In order to enhance the block features and gradient flow, we densely connect the intra dense blocks to further reuse the blocks' information.
The input for each block is the concatenation of all preceding blocks and the output of each block passes on to all the subsequent blocks.
Supposing we have $B$ intra dense blocks.
The input of the $B_{th}$ intra-dense block can be formulated as:
\begin{equation}
L_{B}=[H_{1},H_{2},...,H_{B-1}] \label{9}
\end{equation}
Here $H_{b}$ denotes the output of the $b_{th}$ intra-dense block. More details about the intra-dense block will be shown in next subsection.
And The output of the densely connected $B$ intra-dense blocks $H$ can be formulated as:
\begin{equation}
H=[H_{1},H_{2},...,H_{B}] \label{10}
\end{equation}
\begin{figure*}[t]
\begin{center}
\includegraphics[width=0.96\textwidth,height=0.20\textheight]{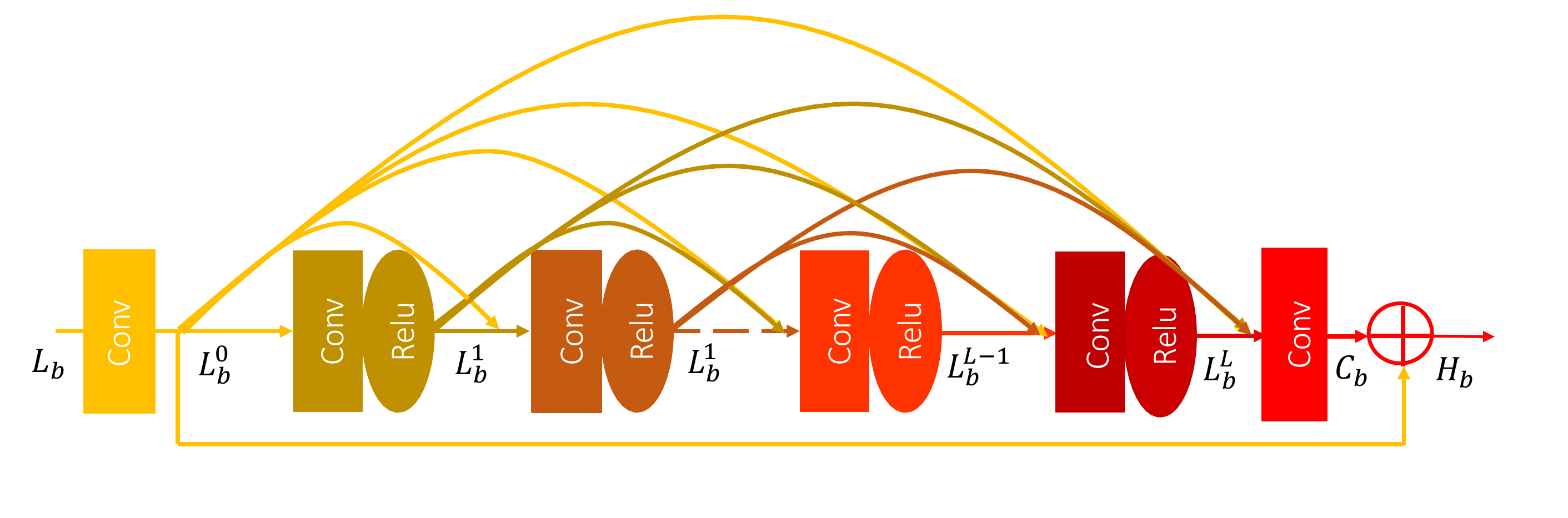}
\caption{Intra dense blocks}
\end{center}
\label{inner}
\end{figure*}
Our inter-block dense connection makes more use of the feature layers and archives better SR performance compared with RDN \cite{b13} and SRDenseNet \cite{b15}. RDN and SRDenseNet use skip connections between blocks to pass only one early block information to the block, but our inter-block dense connection preserves and passes on all early blocks to the block.
We also demonstrate the effectiveness of inter-block dense connection in experiment section, and the results indicate that inter-block dense connection is crucial for image SR.
\subsubsection{Global compression layer}
After producing high level features, we compress the concatenation of all block features into $n_{r}$ channel features with one compression layer. We use one $1\times1$ convolutional layer as the compression layer.
\begin{equation}
T=Conv(1,n_{r})(H)  \label{11}
\end{equation}
Here T denotes the output of the global compression layer.
\subsubsection{Global skip connection}
We add the global skip connection between the input $L^{0}$ and the output $T$ of inter-block dense net.
\begin{equation}
G=T+L^{0} \label{12}
\end{equation}
Here, G denotes the output of global skip connection.
Since the input and output features are highly correlated in SR. Adding a global skip connection that bypasses the input features with an identity function can help gradient flow and mitigate the training of network.
\subsection{Intra-dense blocks}\label{AA}
The intra-dense block is presented in Fig. 4, which contains input compression layer, densely connected layers, output compression layer and the skip connection. Here, let $conv_{b}^{i}(f,n)$ be the $i_{th}$ convolutional layer in the $b_{th}$ intra-dense block, where f is the filter size and n is the number of filters.
\subsubsection{Input compression layer}
Since the input of the $b_{th}$ intra-dense block $L_{b}$ is the concatenation of all the preceding blocks, as the block number increases, the dimension of $L_{b}$ becomes very large. In order to keep the model compact, the input compression layer is used to reduce the feature dimension into $n_{r}$ before entering the dense layers.

We denote the result of input compression layer in the $b_{th}$ intra-dense block as $L_{b}^{0}$ and it can be formulated as:
\begin{equation}
L_{b}^{0}=conv_{b}^{0}(1,n_{r})(L_{b}) \label{13}
\end{equation}
\subsubsection{Densely connected layers}
After reducing the feature vector dimension, a set of densely connected layers are used to learn the high-level features.
The $i_{th}$ layer receives the features of all the preceding layers as input.
We assume the densely connected layers have $L$ layers and each convolution layer is followed by an ReLU function.
Let $L_{b}^{i}$ be the $i_{th}$ layer output in the $b_{th}$ intra dense block. The layer can then be expressed as:
\begin{equation}
L_{b}^{i}=max(0, conv_{b}^{i}(3,n_{g})([L_{b}^{0}, L_{b}^{1},..., L_{b}^{i-1}])) \label{14}
\end{equation}
Here $[L_{b}^{0}, L_{b}^{1},..., L_{b}^{i-1}]$ refers to the concatenation of the features produced in all preceding layers of the $i_{th}$ layer in the $b_{th}$ intra dense block.
Since each layer generates $n_{g}$ dimension features and passes on the $n_{g}$ dimension features to all the subsequent layers, the output of dense connected layers will have $n_{r}+n_{g}\times L$ features.
\subsubsection{Output compression layer}
In order to make the model more compact, we need to reduce feature dimension after densely connected layers. Inspired by the bottle neck structure \cite{b31},
We add one convolutional layer $conv_{b}(1,n_{r})$ as the output compression layer after the densely connected layers to compress the feature dimension from $n_{r}+n_{g}\times L$ to $n_{r}$.
\begin{equation}
C_{b}=conv_{b}^{L+1}(1,n_{r})([L_{b}^{0}, L_{b}^{1},...,L_{b}^{L}]) \label{15}
\end{equation}
Here $C_{b}$ denotes the features of output compression layer in the $b_{th}$ intra-dense block. The reason for adding compression layer is that as model gets deeper and $n_{g}$ becomes larger, the deep densely connected layers without a compression layer are more likely to overfitting and will consume a larger number of parameters without a better performance than the model with the compression layer \cite{b26}. In experiment section, we discuss the importance of compression layer in our model.
\subsubsection{Local skip connection}
Local skip connection is added between the input and output compression layer to help gradient flow during the training of network.
Therefore the output of intra-block dense blocks $H_{b}$ is:
\begin{equation}
H_{b}=C_{b}+L_{b}^{0} \label{16}
\end{equation}

It should be noted that our intra dense block is different from the dense blocks used in DenseNet \cite{b26}, SRDenseNet \cite{b15}, and RDN \cite{b13}. As shown in Fig. 2, compared with the dense block used in DenseNet and SRDenseNet, our intra-dense block is more suitable for SR tasks, by removing BN module to keep the image details and adding compression layer to allow wider and deeper network design. Because of our inter-block dense connection, we add an input compression layer in the intra-dense block to develop the network which is different from the residual dense block in RDN.
\section{Experiment}
\subsection{Implementation and training details}
In the proposed networks, the convolutional layers in the dense connected blocks, transition layer and reconstruction layer are $3 \times3 $ filter size convolutional layers with one padding and one striding.
The feature vector of the extraction layer, transition layer and deconvolutional layer is $n_{r}=64$ dimension.
The base model structure has 16 intra-dense blocks and each block has 8 dense layers. Here we set the feature dimension in each dense connected layer as $n_{g}=n_{r}$.
In the upsampling sub-network of DBDN, for $2 \times $ augmentation, we use a $6 \times6 $ deconvolutional layer with two striding and two padding.
Then, for $3 \times$ augmentation, we use a $9 \times 9$ deconvolutional layer with three striding and three padding.
Finally, for $4 \times$ augmentation, we use two successive $6 \times 6$ deconvolutional layers with two striding and two padding.
In the upsampling sub-network of DBDN+, we use a $Conv(3,a^{2}*n_{r})$ convolutional layer followed by a pixel shuffle layer for $a \times$ augmentation, (a=2,3,4).
We use the same training datasets as EDSR \cite{b14}, the images from Flicker.
To generate the LR image, we downscale the HR images using the bicubic interpolation with scale factors of $2 \times, 3 \times$ and $4 \times$.
In training batch, we use a batch size of 16 HR patches with the size of $96 \times 96$ as the targets and the corresponding LR patches with the size corresponding to the scale factors as inputs.
We randomly augment the patches by flipping and rotating before training.
To keep the image details, instead of transforming the RGB patches into a YCbCr space and only training the Y-channel image information, we use the 3-channel image information from the RGB for training.
The entire network is optimized by Adam \cite{b18} with L1 loss by setting $\beta_{1}=0.9$, $\beta_{2}=0.999$, and $\epsilon=10^{-8}$.
The learning rate is initially set to $10^{-4}$ and halved at every $2\times10^{5}$ minibatch updates for $10^{6}$ total minibatch updates.
\makeatletter
\patchcmd{\@makecaption}
  {\scshape}
  {}
  {}
  {}
\makeatother
\begin{table*}[t]
\caption{Public benchmark test results and Manga109 results (PSNR(dB)/SSIM). Red indicates the best performance and blue indicates the second best.}
\centering

\makebox[1\textwidth][c]{%
\resizebox{0.95\textwidth}{!}{
\begin{tabular}{*{12}{c}}
\hline
\textbf{Datasets} & \textbf{Scale} & \textbf{Bicubic} & \textbf{VDSR \cite{b7}} & \textbf{LapSRN \cite{b16}} & \textbf{DRRN \cite{b6}}  & \textbf{SRDensenet \cite{b15}} & \textbf{EDSR \cite{b14}} &  \textbf{RDN \cite{b13}} & \textbf{DBPN \cite{b32}} & \textbf{DBDN(ours)} & \textbf{DBDN+(ours)}\\
\hline\hline
\multirow{3}{4em}{\textbf{Set5}} & \textbf{$2\times$} & 33.66/0.9929 & 37.53/0.9587 & 37.52/0.9591 & 37.74/0.9591  & -/- & 38.11/0.9601 & 38.24/0.9614 & 38.09/0.9600 & \color{blue}{\textbf{38.30/0.9617}} & \color{red}{\textbf{38.35/0.9618}}\\
& \textbf{$3\times$} & 30.39/0.8682 & 33.66/0.9124 & 33.82/0.9227 & 34.03/0.9244 & -/- & 34.65/0.9282 & 34.71/0.9296 & -/- & \color{blue}{\textbf{34.76/0.9299}} & \color{red}{\textbf{34.83/0.9303}} \\
& \textbf{$4\times$} & 28.42/0.8104 &  31.35/0.8838 & 31.54/0.8855 & 31.68/0.8888 & 32.02/0.8934 & 32.46/0.8968 & 32.47/0.8990 & 32.47/0.8980 & \color{blue}{\textbf{32.54/0.8991}} & \color{red}{\textbf{32.70/0.9006}}\\

\hline\hline
\multirow{3}{4em}{\textbf{Set14}} & \textbf{$2\times$} & 30.24/0.8688 & 33.03/0.9124 & 33.08/0.9130 & 33.23/0.9136  & -/- & 33.92/0.9195 & 34.01/0.9212 & 33.85/0.9190 & \color{blue}{\textbf{34.20/0.9224}} & \color{red}{\textbf{34.34/0.9239}} \\
& \textbf{$3\times$} & 27.55/0.7742 & 29.28/0.8209 & 29.77/0.8314 & 29.96/0.8349 & -/- & 30.52/0.8462 & 30.57/0.8468 & -/- & \color{blue}{\textbf{30.63/0.8478}} & \color{red}{\textbf{30.75/0.8495}}\\
& \textbf{$4\times$} & 26.00/0.7027 &  28.01/0.7674 & 28.19/0.7720 & 28.21/0.7721 & 28.50/0.7782 & 28.80/0.7876 & 28.81/0.7871 & 28.82/0.7860 & \color{blue}{\textbf{28.89/0.7890}} & \color{red}{\textbf{29.00/0.7908}} \\
\hline\hline
\multirow{3}{4em}{\textbf{BSD100}} & \textbf{$2\times$} & 29.56/0.8431 & 31.90/0.8960 & 31.80/0.8950 & 32.05/0.8973 & -/- & 32.32/0.9013 & 32.34/0.9017 & 32.27/0.9000 & \color{blue}{\textbf{32.39/0.9022}} & \color{red}{\textbf{32.45/0.9028}}\\
& \textbf{$3\times$} & 27.21/0.7385 & 28.82/0.7976 & 28.82/0.7973 & 28.95/0.8004 & -/- & 29.25/0.8093 & 29.26/0.8093 & -/- & \color{blue}{\textbf{29.31/0.8104}} & \color{red}{\textbf{29.37/0.8112}}\\
& \textbf{$4\times$} & 25.96/0.6675 & 27.29/0.7251 & 27.32/0.7280 & 27.38/0.7284 & 27.53/0.7337 & 27.71/0.7420 & 27.72/0.7418 &  27.72/0.7400 & \color{blue}{\textbf{27.76/0.7426}} & \color{red}{\textbf{27.84/0.7446}}\\
\hline\hline
\multirow{3}{4em}{\textbf{Urban100}} & \textbf{$2\times$} & 26.88/0.8403 & 30.76/0.8946 & 30.41/0.9101 & 31.23/0.9188 & -/- & 32.93/0.9351 & 32.84/0.9347 & 32.51/0.9321 & \color{blue}{\textbf{32.98/0.9364}} & \color{red}{\textbf{33.36/0.9389}}\\
& \textbf{$3\times$} & 24.46/0.7349 & 26.24/0.7989 & 27.14/0.8272 & 27.53/0.8378 & -/- & 28.80/0.8653 & 28.79/0.8655 & -/- & \color{blue}{\textbf{28.96/0.8682}} & \color{red}{\textbf{29.17/0.8715}}\\
& \textbf{$4\times$} & 23.14/0.6577 & 25.18/0.7524 & 25.21/0.7553 & 25.44/0.7638 & 26.05/0.7819 & 26.64/0.8033 & 26.61/0.8028 & 26.38/0.7945 & \color{blue}{\textbf{26.70/0.8050}} & \color{red}{\textbf{27.00/0.8117}}\\
\hline\hline
\multirow{3}{4em}{\textbf{Manga109}} & \textbf{$2\times$} & 30.80/0.9339 & 37.16/0.9739 & 37.27/0.9740 & 37.60/0.9736 & -/- & 38.96/0.9769 & 39.18/0.9780 & 38.89/0.9775 & \color{blue}{\textbf{39.46/0.9788}} & \color{red}{\textbf{39.65/0.9793}}\\
& \textbf{$3\times$} & 26.95/0.8556 & 31.48/0.9317 & 32.19/0.9334 & 32.42/0.9359 & -/- & 34.17/0.9473 & 34.13/0.9484 & -/- & \color{blue}{\textbf{34.46/0.9498}} & \color{red}{\textbf{34.80/0.9512}}\\
& \textbf{$4\times$} & 24.89/0.7866 & 27.82/0.8856 & 29.09/0.8893 & 29.18/0.8914 & 27.83/0.8782 & 31.11/0.9148 & 31.00/0.9151 & 30.91/0.9137 &\color{blue}{\textbf{31.23/0.9169}} & \color{red}{\textbf{31.68/0.9198}}\\
\hline
%\multicolumn{4}{l}{$^{\mathrm{a}}$Sample of a Table footnote.}
\end{tabular}%
}
}

\label{tab1}

\end{table*}

\begin{figure*}[t]
\begin{center}
\includegraphics[width=0.96\textwidth,height=0.29\textheight]{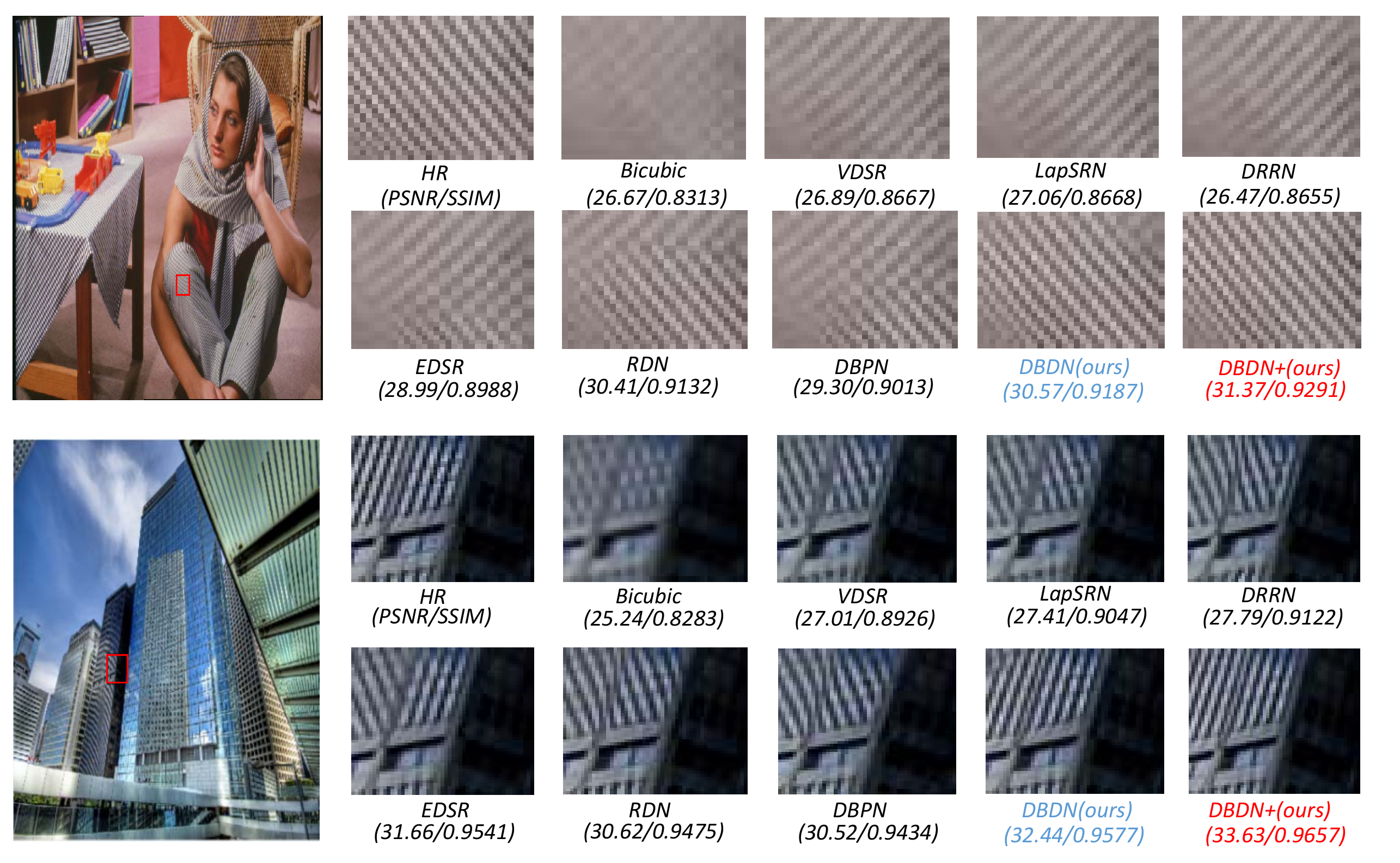}
\caption{Qualitative comparison of our model with other works on $\times2$ super-resolution. Red indicates the best performance and blue indicates the second best.}
\end{center}
\label{feature2}
\end{figure*}
\begin{figure*}[t]
\begin{center}
\includegraphics[width=0.96\textwidth,height=0.30\textheight]{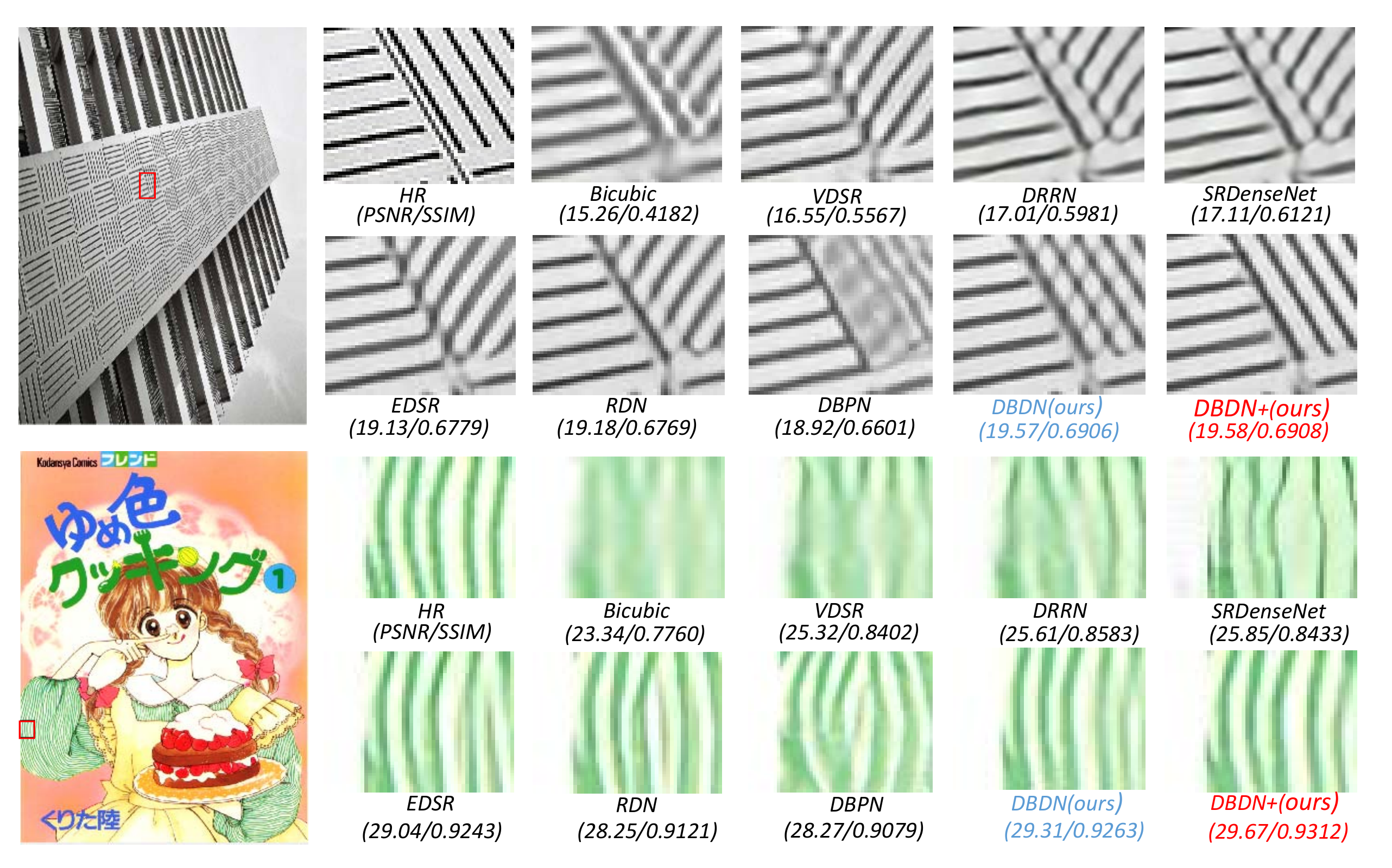}
\caption{Qualitative comparison of our model with other works on $\times4$ super-resolution. Red indicates the best performance and blue indicates the second best.}
\end{center}
\label{feature4}
\end{figure*}

\subsection{Comparison with state-of-art models}
To confirm the ability of the proposed network, we performed several experiments and analyses.
We compared our network with seven state-of-the-art image SR methods: VDSR \cite{b7}, LapSRN \cite{b16}, DRRN \cite{b6}, SRDenseNet \cite{b15}, EDSR \cite{b14}, RDN \cite{b13} and DBPN \cite{b32}.
We carried out the test experiments using five datasets: Set5 \cite{b27}, Set14 \cite{b28}, BSD100 \cite{b29}, Urban100 \cite{b30}, and Manga109 \cite{b17}.
Set5, Set14 and BSD100 are about nature scenes.
Urban100 contains urban building structures and Manga109 is a dataset of Japanese manga.
Table.~\ref{tab1} shows the quantitative results comparisons for $2 \times, 3 \times, 4 \times$ SR.
For comparison, we measure PSNR and SSIM \cite{b10} on the Y-channel and ignore the same amount of pixels as scales from the border.
Note that higher PSNR and SSIM values indicate better quality.
Our methods qualitatively outperform other CNN models with all scale factors in PSNR and SSIM. Compared to the light version CNN networks, DBDN outperforms them more than 1 dB in PSNR; Compared with the recent heavy version CNN networks, DBDN excels EDSR, RDN and DBPN about 0.1 dB in PSNR.
Specifically, for the Manga109 test dataset, our models exhibit significant improvements compared with the other state-of-art methods. Furthermore, DBDN plus surpasses DBDN about 0.1 dB.

We also provide visual comparison results as qualitative comparisons.
Fig. 5 shows the visual comparisons on the $2\times$ scale. For image 'barbara', all our methods can recover sharper and clearer pattern that are subjectively closer to the ground truth, while most of the compared methods generate blurred or biased cloth pattern. Similarly, for image 'img061' in the Urban100 dataset, all our methods can accurately recover the building structures. However, all the compared methods produce biased building lines. Fig. 6 illustrates the qualitative analysis on the $4\times$ scale. Our methods suppress the blurring artifacts, recover patterns closer to the ground truths and exhibit better-looking SR outputs compared with the previous methods.
This comparison demonstrates the effectiveness of our methods for image SR tasks.
\subsection{Model analysis}

\subsubsection{Number of parameters}
To demonstrate the compactness of our model, we compare the model performance and network parameters of our model with existing deep networks for image SR in Fig. 7.
Our model shows the trade-off between the parameter demands and performance.
Since VDSR \cite{b7}, DRRN \cite{b6}, LapSRN \cite{b16}and SRDenseNet \cite{b15} are all light version networks, they all visibly concede the performance for the model parameter numbers.
Although DBDN has more parameters than DBPN \cite{b32}, DBDN has about 0.1dB higher performance than DBPN on Set5 for $4\times$ enlargement.
Compared with EDSR \cite{b14}, which is one of the previous best performances, our DBDN network achieves the higher performance with about $58 \%$ fewer parameters.
Compared with RDN \cite{b13}, DBDN achieves about 0.1 dB higher performance on Set5 for $4\times$ enlargement with the same parameter numbers as RDN. Moreover, our DBDN+ outperforms all the other methods by a large margin with the same parameter numbers as DBDN.
\begin{figure}[h]
\includegraphics[width=0.45\textwidth,height=0.20\textheight]{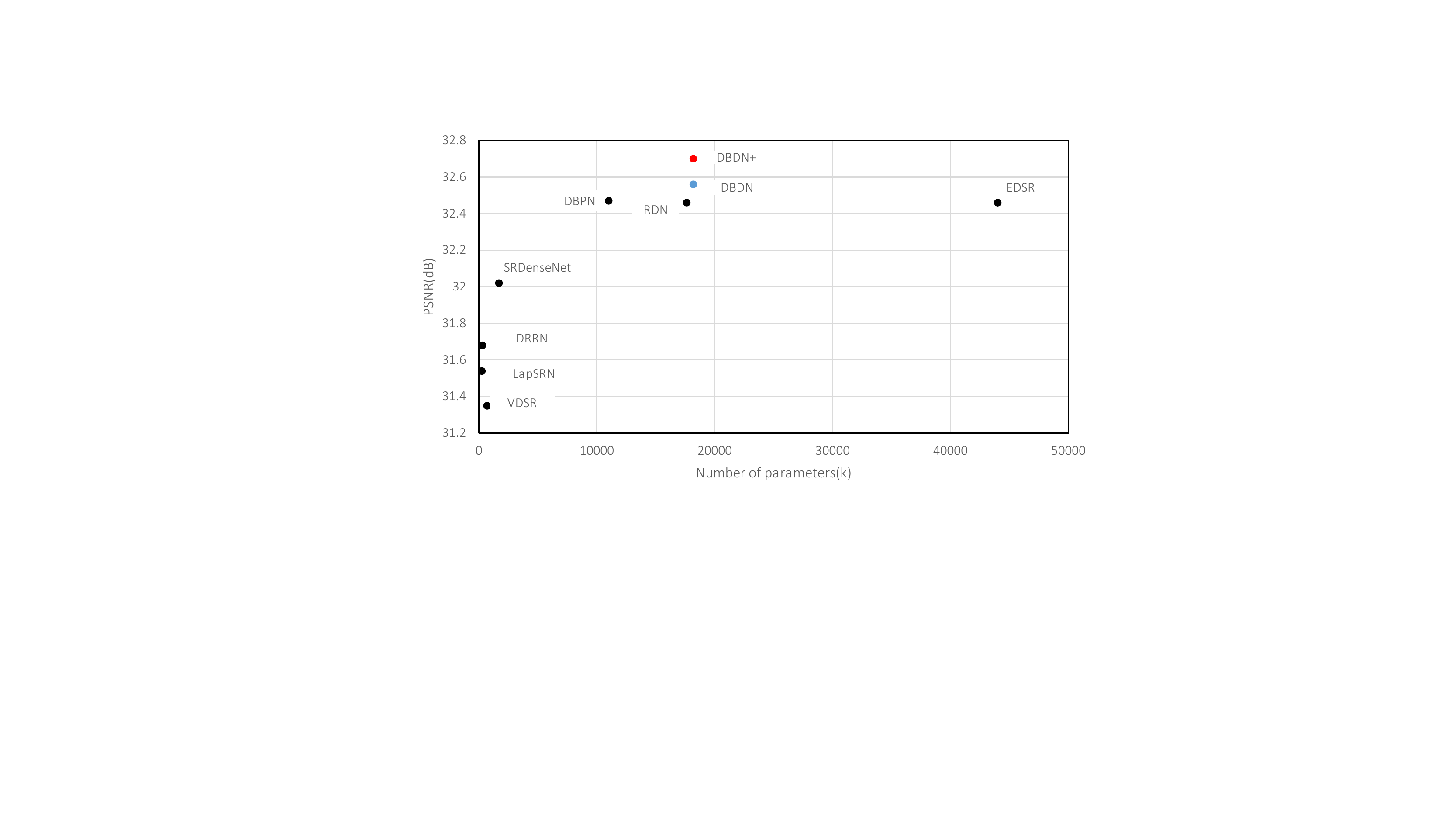}
\caption{Performance vs number of parameters. The results are evaluated with Set5 for $4\times$ enlargement. Red indicates the best performance and blue indicates the second best.}
\label{compare}
\end{figure}
\subsubsection{Ablation investigation}
In this section, we will analyze the effects of different network modules on the model performance. Since skip connection has been discussed in many deep learning methods \cite{b13}\cite{b6}\cite{b15}, we focus on the effects of the compression layer and the inter-block dense connection in our model.
\begin{figure}[h]
\includegraphics[width=0.45\textwidth,height=0.20\textheight]{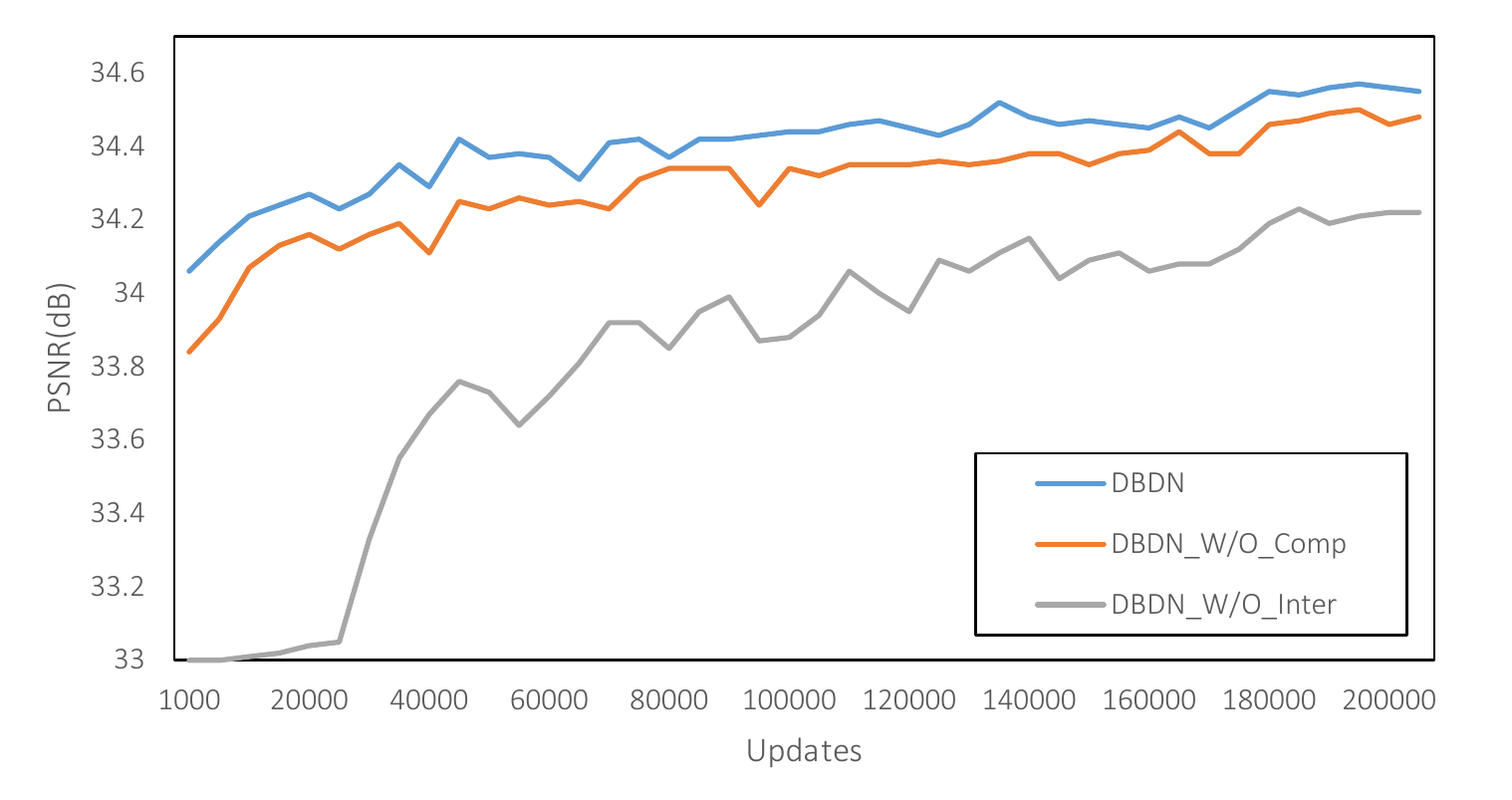}
\caption{Discussion about transition layer, skip connection and inter-block dense connectivity in DBDN. The results are evaluated with Set5 for $3\times$ enlargement in 200 epochs}
\label{analysis1}
\end{figure}
To demonstrate the effectiveness of the compression layer, we create a network without compression layer between the blocks.
In order to form a fair comparison with the baseline model, the depth and number of parameters are kept the same for both methods. Therefore we set the model to have 128 convolutional layers in the block, each layer with 16 filters of the size of $3 \times 3$ and only one dense block to construct the total network. We denote this model as DBDN-W/O-Comp.
To demonstrate the effectiveness of the inter-block dense connection, we chained the intra-dense blocks into line to construct a DBDN-W/O-Inter for learning the mapping function between the LR and HR image. The parameter setting of DBDN-W/O-Inter is also the same as DBDN.
We visualize the convergence process of these models. As we can see from Fig. 8, The models without the compression layers and the inter-block dense connection suffer performance drop in the training, even when the models have the same condition of parameters . Especially for the inter-block dense connection, the best performance of DBDN on Set5 for $3 \times$ augmentation in 200 epoches is 0.6 dB higher than DBDN-W/O-Inter. Therefore, the inter-block dense connection is crucial for image SR performance.
\section{Conclusion}
We have proposed a DBDN for single image SR. Unlike the previous methods that only reuse several feature layers in a local dense block by using a dense connection, our proposed network extends previous intra-block dense connection approaches by including inter-block dense connections. The bi-dense connection structure helps the gradient and feature flows between the layers to archive better performance. The proposed method outperforms the state-of-art methods by a considerable margin on five standard benchmark datasets in terms of PSNR and SSIM. The noticeable improvement can also visually be found in the reconstruction results. We also demonstrate that the different modules in our network improve the performance at different levels, and the inter-block dense connection has key contribution to our outstanding performance. Further work will focus on training bi-dense network with perceptual loss.


\begin{thebibliography}{00}
\bibitem{b1} C. Dong, C. Loychen, H. Kaiming, and T. Xiaoou. ``Learning a deep convolutional network for image super-resolution,'' European conference on computer vision. pp. 184-199, September 2014.
\bibitem{b2} K. Kwangin, and K. Younghee. ``Learning a deep convolutional network for image super-resolution,'' IEEE Trans. Pattern Anal. Mach. Intell. vol. 32, pp. 1127--1133, 2010.
\bibitem{b3} K. Robert. ``Cubic convolution interpolation for digital image processing,'' IEEE Trans. Signal Process. vol. 29, pp. 1153--1160, 1981.
\bibitem{b4} W. Shi, J. Caballero , and C. Ledig et al. ``Cardiac image super-resolution with global correspondence using multi-atlas patchmatch,'' International Conference on Medical Image Computing and Computer-Assisted Intervention. Berlin. Heidelberg, pp. 9-16, 2013.
\bibitem{b5} Y. Tai, J. Yang, and X. Liu. ``Image super-resolution via deep recursive residual network,'' IEEE Conference on Computer Vision and Pattern Recognition. 2017.
\bibitem{b6} K. Jiwon , K. Leejung, and M. Leekyoung. ``Deeply-recursive convolutional network for image super-resolution,'' IEEE Conference on Computer Vision and Pattern Recognition. pp. 1637--1645. 2016.
\bibitem{b7} K. Jiwon , K. Leejung, and M. Leekyoung. ``Accurate image super-resolution using very deep convolutional networks,'' IEEE Conference on Computer Vision and Pattern Recognition. pp. 1646--1654. 2016.
\bibitem{b8} H. Kaiming, Zhang. Xiangyu, Ren. Shaoqing, and S. Jian. ``Deep residual learning for image recognition,'' IEEE conference on computer vision and pattern recognition. pp. 770--778. 2016.
\bibitem{b9} D. Chao, L. Chenchange, H. Kaiming, and T. Xiaoou. ``Image super-resolution using deep convolutional networks,'' IEEE Trans. Pattern Anal. Mach. Intell. vol. 38, pp. 295--307, 2016.
\bibitem{b10} Z. Wang, C. BovikAlan, R.SheikhHamid, and P. SimoncelliEero. ``Image quality assessment: from error visibility to structural similarity,'' IEEE Trans. Signal Process. vol. 13, pp. 600--612, 2004.
\bibitem{b11} S. Ioffe, and C. Szegedy. ``Batch normalization: Accelerating deep network training by reducing internal covariate shift,'' International Conference on Machine Learning. pp. 448--456, 2015.
\bibitem{b12} V. Nair, and E. HintonGeoffrey. ``Rectified linear units improve restricted boltzmann machines,'' International conference on machine learning. pp. 807--814, 2010.
\bibitem{b13} Z. Yulun, T. Yapeng, Y. Kong, Z. Bineng, and F. Yun. ``Residual dense network for image super-resolution,'' IEEE Conference on Computer Vision and Pattern Recognition. 2018.
\bibitem{b14} B. Lim, S. Son, H. Kim, S. Nahet, and K. MuLee. ``Enhanced deep residual networks for single image super-resolution,'' IEEE conference on computer vision and pattern recognition workshops. vol. 1, p. 4, 2017.
\bibitem{b15} T. Tong, G. Li, X. Liu, and Q. Gao. ``Image super-resolution using dense skip connections,'' IEEE International Conference on Computer Vision. 2017.
\bibitem{b16} L. WeiSheng, H. JiaBin, A. Narendra, and Y. MingHsuan. ``Deep Laplacian Pyramid Networks for Fast and Accurate Super-Resolution,'' IEEE Conference on Computer Vision and Pattern Recognition. Vol. 2. No. 3. 2017.
\bibitem{b17} Y. Matsui, K. Ito, Y. Aramaki, A. Fujimoto, T. Ogawa, T. Yamasaki, and K.Aizawa. ``Sketch-based manga retrieval using manga109 dataset,'' Multimedia Tools and Applications. May 2017.
\bibitem{b18} D. Kingma, and J. Ba. ``Adam: A method for stochastic optimization,'' ICLR. May 2016.
\bibitem{b19} W. Shi, J. Caballero, F. Husz¨¢r, J. Totz, A.P. Aitken, R. Bishop, D. Rueckert, and Z. Wang. ``Real-time single image and video super-resolution using an efficient sub-pixel convolutional neural network,'' IEEE Conference on Computer Vision and Pattern Recognition. pp. 1874-1883, 2016.
\bibitem{b20} C. Dong, L. ChenChange, and T. Xiaoou. ``Accelerating the super-resolution convolutional neural network,'' European Conference on Computer Vision. pp. 391-407, 2016.
\bibitem{b21} W. W. Zou, and P. C. Yuen. ``Very low resolution face recognition problem,'' IEEE Trans. Image Processing. 2012.
\bibitem{b22} D. Martin, C. Fowlkes, D. Tal, and J. Malik. ``A database of human segmented natural images and its application to evaluating segmentation algorithms and measuring ecological statistics,'' IEEE International Conference on Computer Vision. pp. 416--423, 2001.
\bibitem{b23} W. M. Thornton, M.P. Atkinson, and D. Holland. ``Sub-pixel mapping of rural land cover objects from fine spatial resolution satellite sensor imagery using super-resolution pixel-swapping,'' International Journal of Remote Sensing. pp. 473--491, 2006.
\bibitem{b24} Y. Jianchao, J. Wright, T.S. Huang, and Y. Ma. ``Image super-resolution via sparse representation,'' IEEE Trans. Image Processing. pp. 2861--2873, 2012.
\bibitem{b25} Y. Tai, Y. Jian, L. Xiaoming, and X. Chunyan. ``Memnet: A persistent memory network for image restoration,'' IEEE Conference on Computer Vision and Pattern Recognition. pp. 4539-4547. 2017.
\bibitem{b26} G. Huang, Z. Liu, L. Van Der Maaten, and K. Q. Weinberger. ``Densely Connected Convolutional Networks,'' IEEE Conference on Computer Vision and Pattern Recognition. 2017.
\bibitem{b27} M. Bevilacqua, A. Roumy, C. Guillemot, and M. Alberi-Morel. ``Low-complexity single-image super-resolution based on nonnegative neighbor embedding,'' 2012.
\bibitem{b28} R. Zeyde, M. Elad, and M. Protter. ``On single image scale-up using sparse-representations,'' International conference on curves and surfaces. pp.711--730. 2010.
\bibitem{b29} R. Timofte, V. De Smet, and L. Van Gool. ``A+: Adjusted anchored neighborhood regression for fast super-resolution,'' Asian Conference on Computer Vision. pp.111--126. 2014.
\bibitem{b30} H. Jia-Bin, S. Abhishek, and A. Narendra. ``Single image super-resolution from transformed self-exemplars,'' IEEE Conference on Computer Vision and Pattern Recognition. pp.5197--5206. 2015.
\bibitem{b31} M. Lin, C. Qiang, and Y. Shuicheng. ``Network in network,'' Neural and Evolutionary Computing. 2013.
\bibitem{b32} H. Muhammad, S. Greg, and U. Norimichi. ``Deep Back-Projection Networks For Super-Resolution,'' IEEE Conference on Computer Vision and Pattern Recognition. 2018.


\end{thebibliography}
\end{document}